\documentclass{ifacconf}

\usepackage{graphicx}      
\usepackage{subfigure}
\usepackage{natbib}        
\begin{document}
\begin{frontmatter}

\title{Non-Autoregressive vs Autoregressive Neural Networks for System Identification\thanksref{footnoteinfo}}

\thanks[footnoteinfo]{This work was partly funded by the German Federal Ministry of Education and Research (FKZ: 16EMO0262).}

\author[First]{Daniel Weber} 
\author[Second]{Clemens Gühmann} 

\address[First]{Technische Universität Berlin, 
   Berlin, 10623 Germany (e-mail: d.weber.1@tu-berlin.de).}
\address[Second]{Technische Universität Berlin, 
   Berlin, 10623 Germany (e-mail: clemens.guehmann@tu-berlin.de)}

\begin{abstract}
The application of neural networks to non-linear dynamic system identification tasks has a long history, which consists mostly of autoregressive approaches. Autoregression, the usage of the model outputs of previous time steps, is a method of transferring a system state between time steps, which is not necessary for modeling dynamic systems with modern neural network structures, such as gated recurrent units (GRUs) and Temporal Convolutional Networks (TCNs). We compare the accuracy and execution performance of autoregressive and non-autoregressive implementations of a GRU and TCN on the simulation task of three publicly available system identification benchmarks. Our results show, that the non-autoregressive neural networks are significantly faster and at least as accurate as their autoregressive counterparts. Comparisons with other state-of-the-art black-box system identification methods show, that our implementation of the non-autoregressive GRU is the best performing neural network-based system identification method, and in the benchmarks without extrapolation, the best performing black-box method.
\end{abstract}


\end{frontmatter}

\section{Introduction}
\begin{figure}
    \centering
    \subfigure[Autoregressive model]{\includegraphics[width=0.23\textwidth]{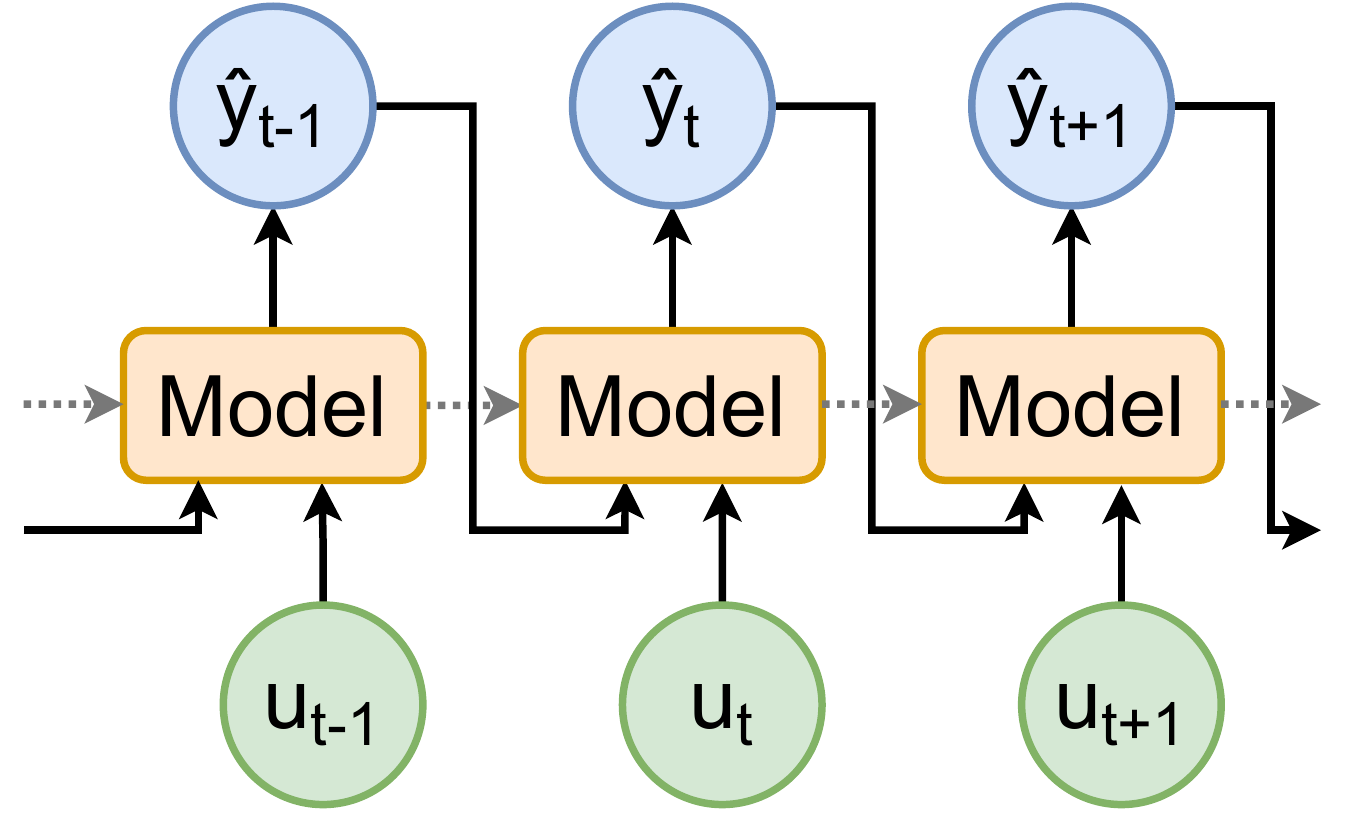}}
    \subfigure[Non-autoregressive model]{\includegraphics[width=0.23\textwidth]{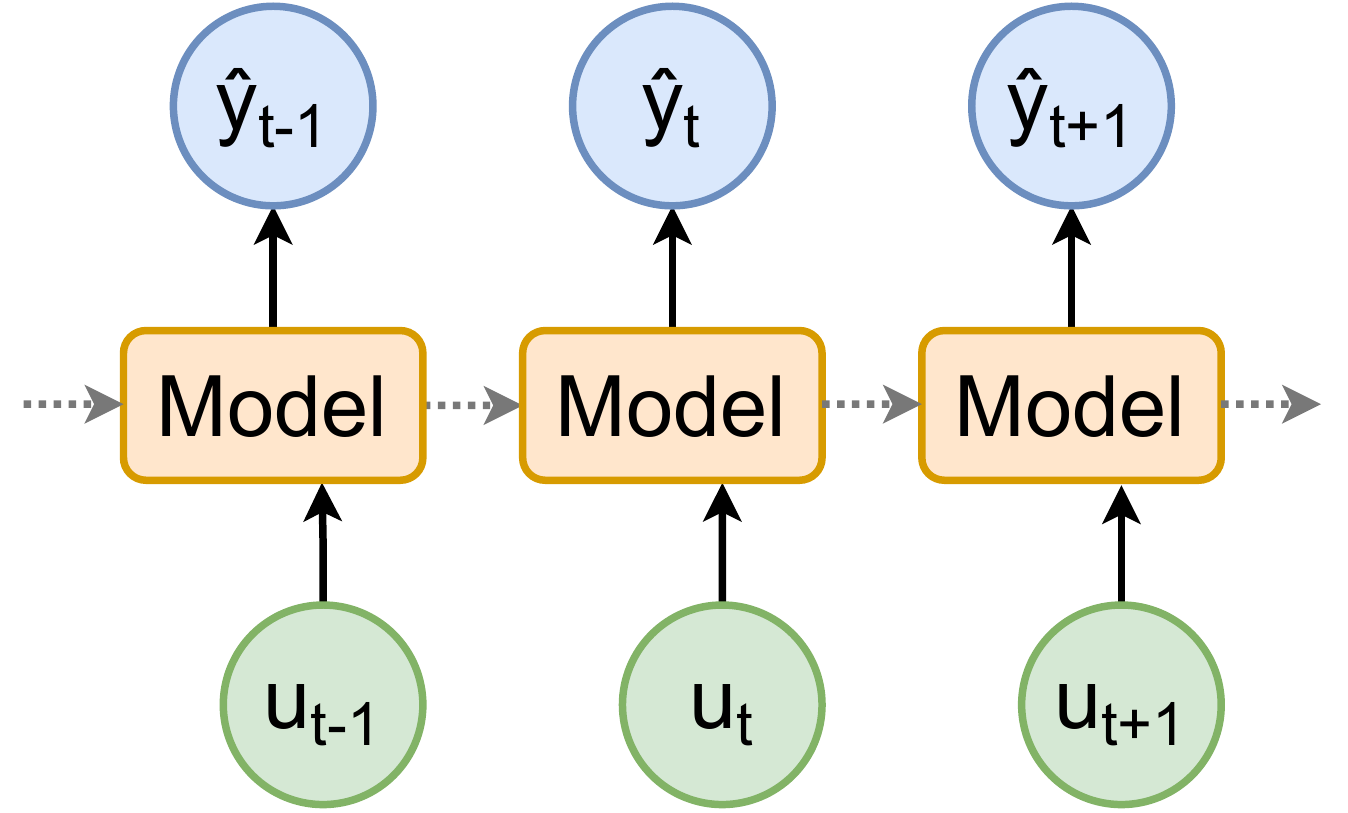}}
    \caption{Signal flow of an autoregressive and a non-autoregressive model with an optional internal state for three time steps.}
    \label{fig:autregression}
\end{figure}

System identification is essential for many tasks, such as system control and sensor fusion. It has a long history, beginning with the identification of linear systems \citep{zadeh_identification_1956}. Most practical relevant systems are nonlinear and dynamic, which led to the development of nonlinear dynamic system identification methods.

Those models may be autoregressive or non-autoregressive, as visualized in Figure \ref{fig:autregression}. The output of an autoregressive model is dependant on the model outputs of previous time steps, which is not the case for a non-autoregressive model. In both cases, the model may have an internal state, which is transferred between each time step. An example for an autoregressive model without an internal state is the nonlinear autoregressive exogenous (NARX) model and for a non-autoregressive model with an internal state is the non-linear state space (NLSS) model. 

The identified models may be used for prediction or simulation. Prediction is the task of estimating a limited amount of time steps ahead with information of the past system outputs. Simulation is the task of estimating the system output only with the inputs. In the present work, we will focus only on the simulation task.

Neural networks have been applied to system identification tasks for a long time. Historically feedforward neural networks are applied autoregressively to model the system dynamics, which is a variant of a NARX model \citep{chen_non-linear_1990}. The success of this approach was limited by the available hardware and the gradient propagation over long sequences. With the recent development of deep learning-based methods in computer vision and natural language processing, more sophisticated software and hardware for the training of neural networks became available.

With this development, the application of neural networks for system identification has become widespread, with a focus on improving upon existing black-box system identification methods. Most current neural network-based system identification methods still are NARX variants with different neural network architectures as nonlinearities. In related work, a multitude of system identification methods that are based on autoregressive neural networks has been proposed using multilayer perceptrons (MLPs) \citep{shi_neural_2019}, cascaded MLPs \citep{ljung_deep_2020}, convolutional neural networks \citep{lopez_nonlinear_2017}, TCNs \citep{andersson_deep_2019}, and recurrent neural networks \citep{kumar_comparative_2019} with promising results.

Although neural networks that are based on convolutional or recurrent layers have inherent capabilities to model dynamic systems, only a limited amount of work omitted the autoregression. It has been shown, that RNNs behave like an NLSS \citep{ljung_deep_2020} and an RNN has been applied non-autoregressively to a synthetic dataset \citep{gonzalez_non-linear_2018}. The authors of the present paper applied non-autoregressive TCNs and gated recurrent units (GRUs) to an inertial measurement-based sensor fusion task, outperforming state-of-the-art domain-specific sensor fusion methods \citep{weber_neural_2020}. 

To the best of our knowledge, it has not yet been analyzed what the differences in the implementation, accuracy, and execution performance of autoregressive and non-autoregressive neural networks are.

The present paper focuses on the comparison of autoregressive and non-autoregressive variants of TCNs and GRUs with the following main contributions:
\begin{itemize}
    \item We describe a workflow for autoregressive and non-autoregressive neural networks for system identification using state-of-the-art deep learning tools and methods.
    \item We find that non-autoregressive neural networks are faster and easier to implement than their autoregressive counterparts and are as least as accurate.
    \item We compare the estimation performance of the proposed neural networks and state-of-the-art black-box system identification methods on three publicly available benchmark datasets.
    \item We find that non-autoregressive gated recurrent units consistently outperform all other neural network-based system identification models and in the benchmarks without extrapolation, all black-box models.
\end{itemize}

\section{Neural Network Implementation}
\label{sec:networks}
In this section, we describe two different neural network architectures, which have the inherent ability to model dynamic systems without autoregressive connections. Furthermore, we describe the corresponding training process with best practices from sequential data processing with neural networks. The two architectures that we consider are TCNs, which process large sequences at once, and GRUs which propagate hidden states over time.

\begin{figure}
    \centering
    \subfigure[TCN-AR]{\includegraphics[width=0.22\textwidth]{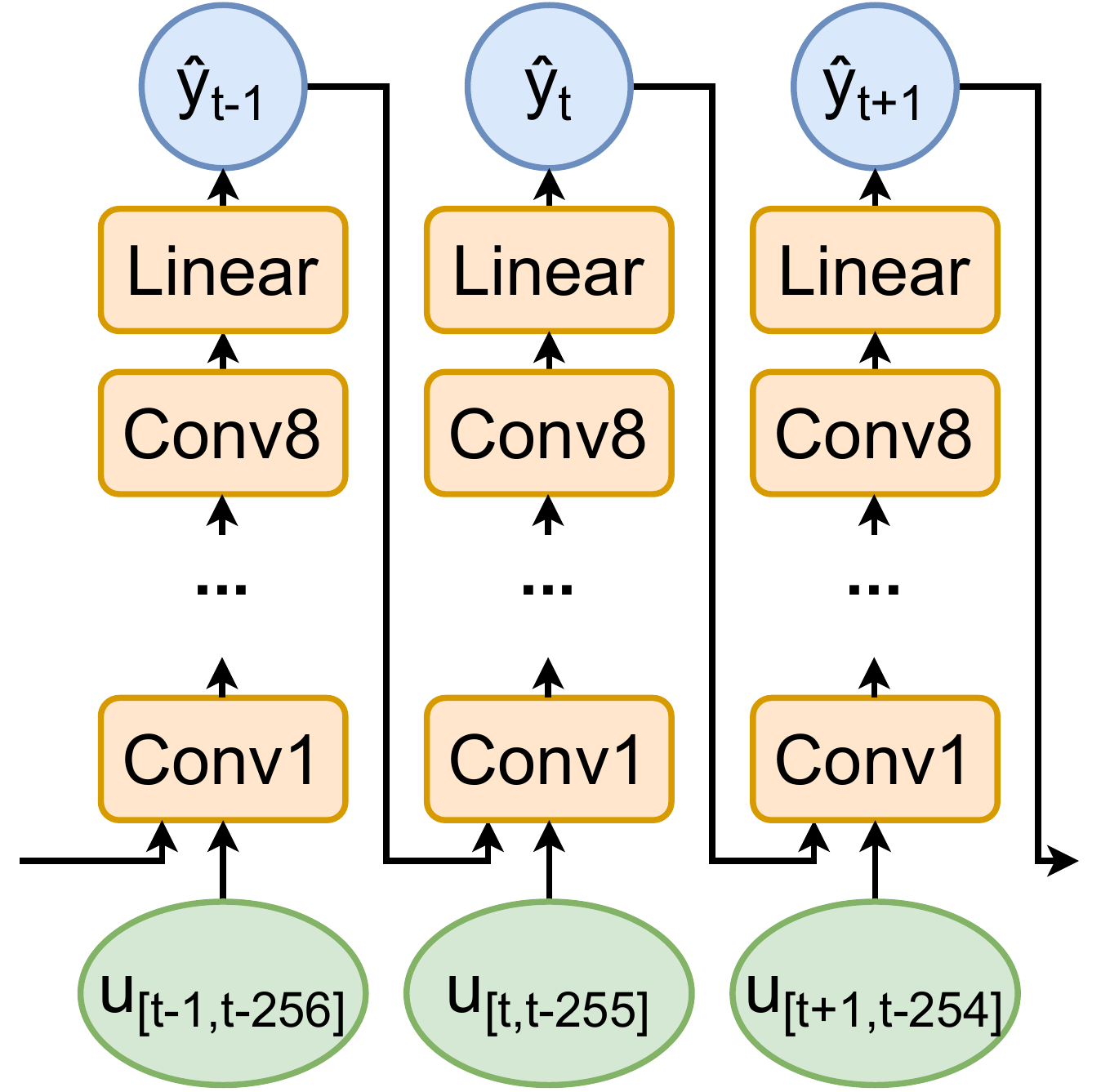}}
    \subfigure[TCN-NAR]{\includegraphics[width=0.22\textwidth]{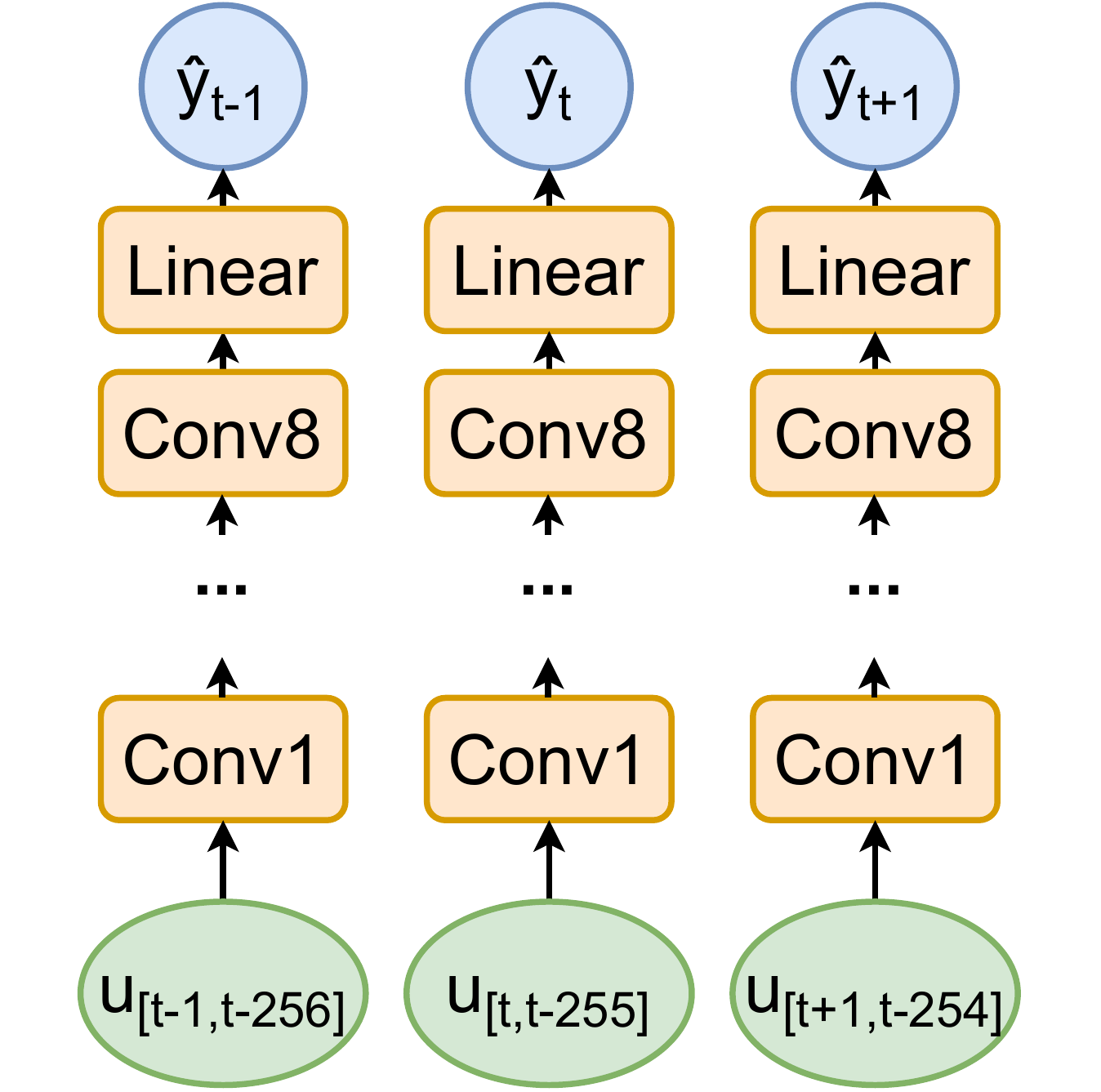}}
    \caption{Signal flow of the autoregressive TCN (TCN-AR) and non-autoregressive TCN (TCN-NAR) for three time steps.}
    \label{fig:tcn}
\end{figure}

TCNs are stateless feed-forward networks that apply 1d-dilated causal convolutions to sequences \citep{andersson_deep_2019}, which are inspired by Wavenet, which is used for raw audio generation \citep{oord_wavenet_2016}. The convolutional layers are stacked on top of each other, which results in a receptive field that describes the number of input samples that are taken into account for the prediction of an output value. The dilation in the convolutions increases the receptive field exponentially with the layer size, enabling the processing of sequences with relations over long periods of time. The main advantage of TCNs is that they are fast in training and inference because of their high parallelizability. The main disadvantage of TCNs is that their ability to model system dynamics is limited by the size of the receptive field, limiting its viability in systems that are influenced by integrators over an indefinite amount of time \citep{weber_neural_2020}. 

\begin{figure}
    \centering
    \subfigure[GRU-AR]{\includegraphics[width=0.22\textwidth]{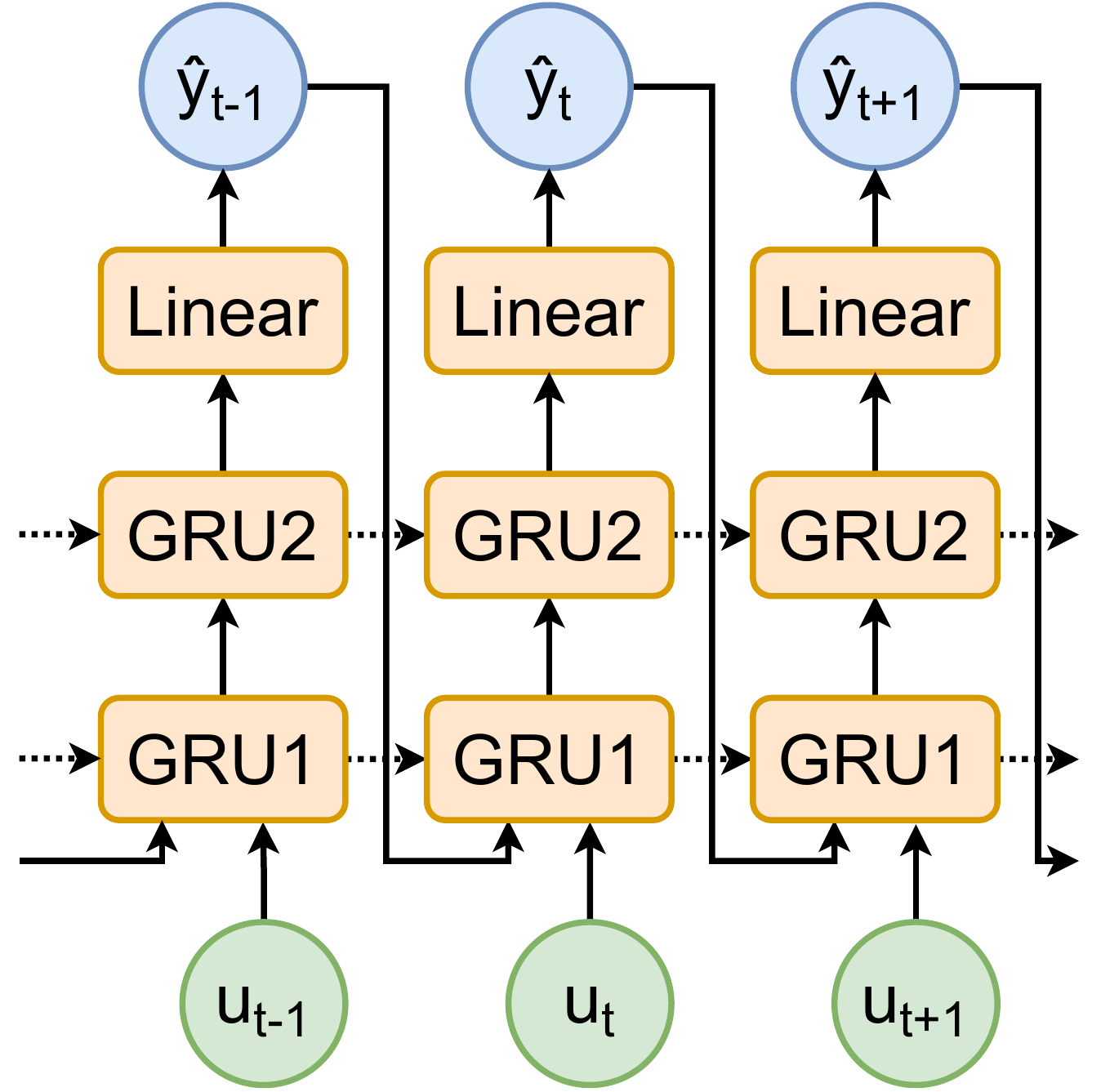}}
    \subfigure[GRU-NAR]{\includegraphics[width=0.22\textwidth]{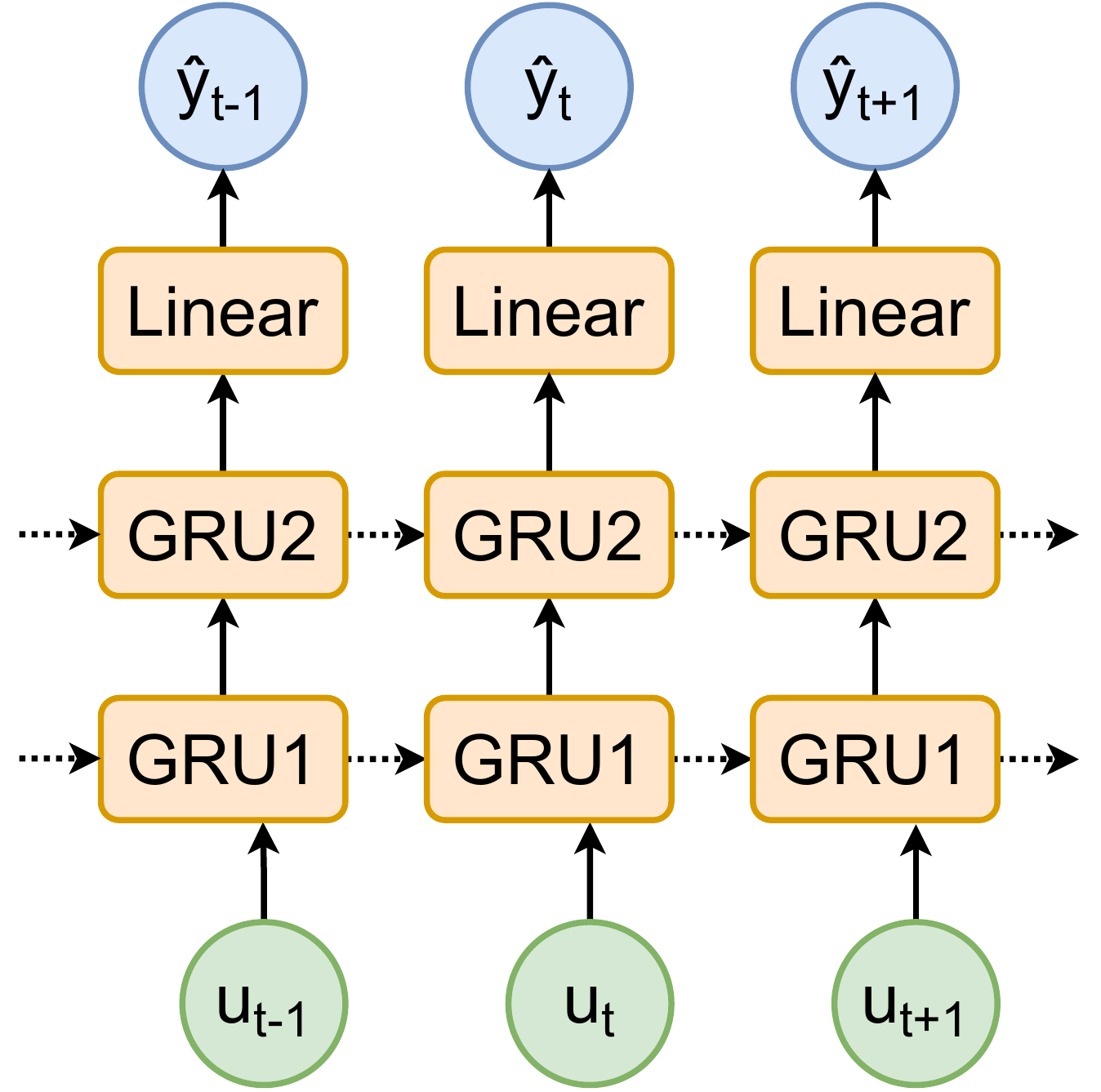}}
    \caption{Signal flow of the autoregressive GRU (GRU-AR) and non-autoregressive GRU (GRU-NAR) for three time steps.}
    \label{fig:gru}
\end{figure}

GRUs are a variant of RNNs, that use recurrent connections in their hidden layers to store a hidden state for each time step. This approach has the advantage, that the state information can be theoretically stored over an indefinite amount of time, which in practice is limited by the vanishing gradient problem \citep{hochreiter_vanishing_2011}. 

The vanishing gradient problem stems from numerical issues on long backpropagation paths in the optimization process. In autoregressive neural networks, the backpropagation path is very long because for every time step the gradient has to propagate from the output through all layers back to the input. In RNNs, the backpropagation path is shorter, but the vanishing gradient problem is still limiting at hundreds of time steps \citep{hochreiter_vanishing_2011}.

The gating mechanism of GRUs alleviates this issue, enabling the network to propagate the hidden state over thousands instead of hundreds of time steps during training. Several regularization methods for RNNs have been proposed, that reduce overfitting and improve generalizability \citep{merity_regularizing_2017}. The main disadvantage of GRUs over TCNs is their sequential nature, which limits its parallelizability, reducing the training speed and especially the inference speed on acceleration hardware.

Both network architectures may be used as autoregressive models, which we refer to as GRU-AR and TCN-AR, and as non-autoregressive models, which we refer to as GRU-NAR and TCN-NAR. The structures of these models are visualized in Figure \ref{fig:tcn} and \ref{fig:gru}. Autoregressive models may be trained using teacher-forcing, with the ground-truth values as input, or in free-running mode, with its own output as input \citep{lamb_professor_2016}. Teacher forcing is faster in training, but only accurate for one-step-ahead prediction tasks, where the model only has to predict the next sample with the values measured in the past. For simulation with neural networks, teacher-forcing is inaccurate, which is why we train the autoregressive models in free-running mode \citep{ribeiro_parallel_2018}.

The naive application of TCN-AR would be very slow and requires an enormous amount of memory during training, because of the number of redundant calculations in its broad receptive field. Because of that, we implemented an optimized caching scheme, which enables us to train TCN-AR in free-running mode in the first place\citep{paine_fast_2016}.

We implemented the training process with FastAI 2, which is a deep learning library that is built upon Pytorch \citep{howard_fastai_2020}. For the optimizer, we use the current state-of-the-art combination of RAdam and Lookahead which proved to be effective in various tasks and requires no learning-rate warm-up phase  \citep{liu_variance_2019}, \citep{zhang_lookahead_2019}. Because we need no warm-up phase, we use cosine-annealing for decreasing the learning rate as soon as the optimizer hits an optimization plateau \citep{loshchilov_sgdr_2017}. We find the maximum learning rate, with which the optimizer begins, with the learning rate finder heuristic \citep{smith_cyclical_2017}.

The datasets used for training consist of multiple measured sequences of different lengths. We extract overlapping windows with varying starting points from the sequences for the generation of the mini-batches, to avoid memorizing the sequences. To process longer sequences, we use truncated backpropagation through time (TBPTT) \citep{tallec_unbiasing_2017}. To apply TBPTT to autoregressive models, not only the hidden state but also the last generated output has to be transferred between two mini-batches. The input signals are standardized to zero mean and a standard deviation of one, which is also applied just-in-time to the autoregressive inputs \citep{ioffe_batch_2015}.

We use the elementwise mean-squared error as the loss function, which is close to the evaluation metric root-mean-square error. Every estimation of dynamic systems that starts with an unknown system state has a transient phase, that describes the time period the estimator needs to converge to a steady-state solution. To minimize the long-term error, we exclude the transition phase from the loss function in the optimization process. With TBPTT, only the first mini-batch of a sequence is affected, because all following mini-batches have a converged system state. TCNs have a receptive field that describes the number of samples that are taken into account for each estimated value. We exclude the estimations that have fewer input values than the receptive field from the gradient propagation because it relies on padded values with no measured information.

The optimization process of the neural network requires a multitude of hyperparameters that have a significant influence on the final performance of the model. Because the hyperparameters span a vast optimization space, it is difficult to find the optimal configuration without an efficient optimization algorithm. For this, we use the Asynchronous Successive Halving Algorithm (ASHA) \citep{li_system_2020}, a state-of-the-art black-box optimizer that combines random-search with early-stopping in an asynchronous environment.

\section{Performance Evaluation}
In this section, we evaluate the accuracy and the simulation time of the neural network models, which we described in Section \ref{sec:networks}, on several publicly available datasets. Additionally, we compare the performance of the neural network models with the results of state-of-the-art system identification methods of related work. All models have been trained and executed on an Nvidia RTX 2080 Ti.

The following datasets are designed specifically as benchmarks for non-linear system identification methods:
\begin{itemize}
    \item Silverbox \citep{wigren_three_2013}: The Silverbox benchmark was generated from an electrical circuit that models a nonlinear progressive spring as an oscillating system. The main challenge of the dataset is the extrapolation part of the test sequence, where the output values are larger than any value in the estimation sequence. 
    \item Wiener-Hammerstein \citep{schoukens_wiener-hammerstein_2009}: The Wiener Hammerstein benchmark was generated by an electrical circuit that models a Wiener-Hammerstein system.
    \item Wiener Hammerstein with Process Noise \citep{schoukens_wiener-hammerstein_2016}: The Wiener-Hammerstein benchmark with process noise was generated by an electrical circuit that models a Wiener-Hammerstein system with additive process noise in the training data. Identifying the process without the process noise is the main challenge of the dataset.
\end{itemize}

All benchmark datasets provide an estimation and a test subset. For the training and hyperparameter optimization of the neural networks, we split the given estimation subsets further in training and validation subsets. For the evaluation of the performance, we use the root-mean-square error (RMSE). We also ignore the output of the first $N$ samples in the transient phase of the simulation with $N$ given in the description of each benchmark dataset.

For every dataset, the hyperparameters for the GRU and the TCN have been optimized. The resulting configuration is used to train an autoregressive and a non-autoregressive variant, that is used for the following evaluation.

\subsection{Autoregressive vs Non-Autoregressive Neural Networks}
\begin{figure}
    \centering
    \subfigure[WH]{\includegraphics[width=0.15\textwidth]{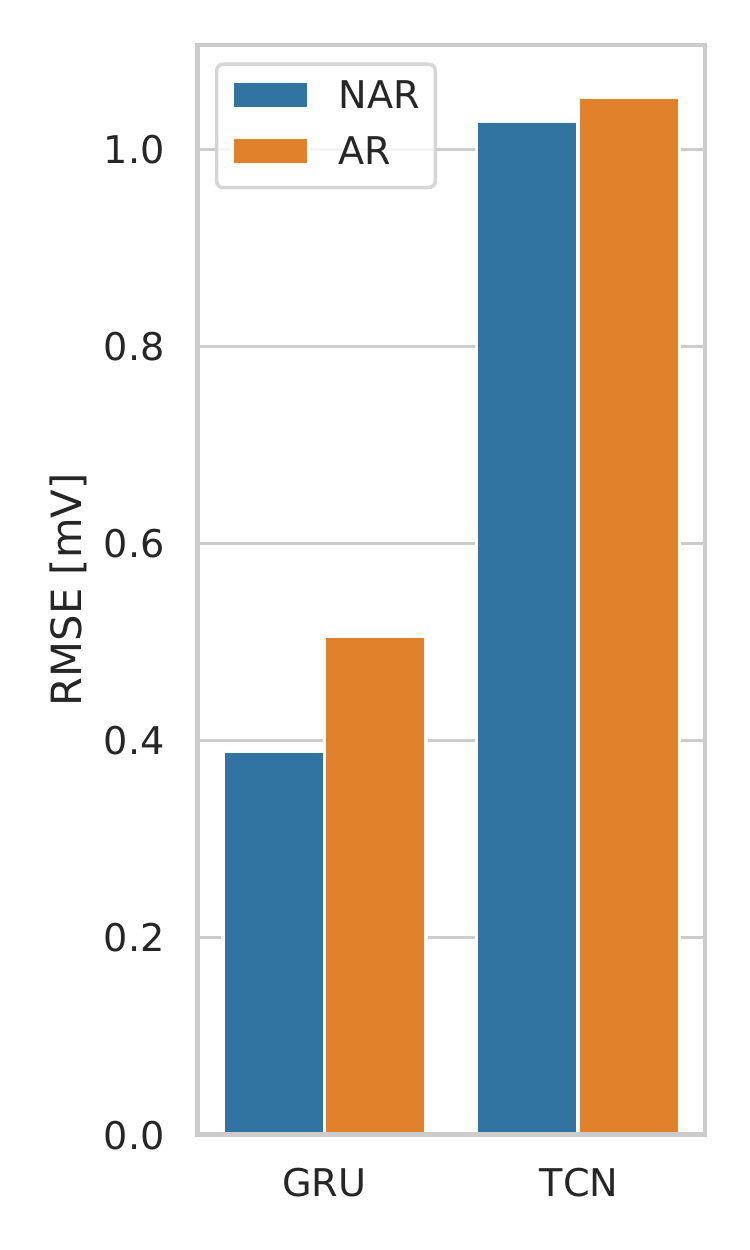}}
    \subfigure[WH-Noise]{\includegraphics[width=0.15\textwidth]{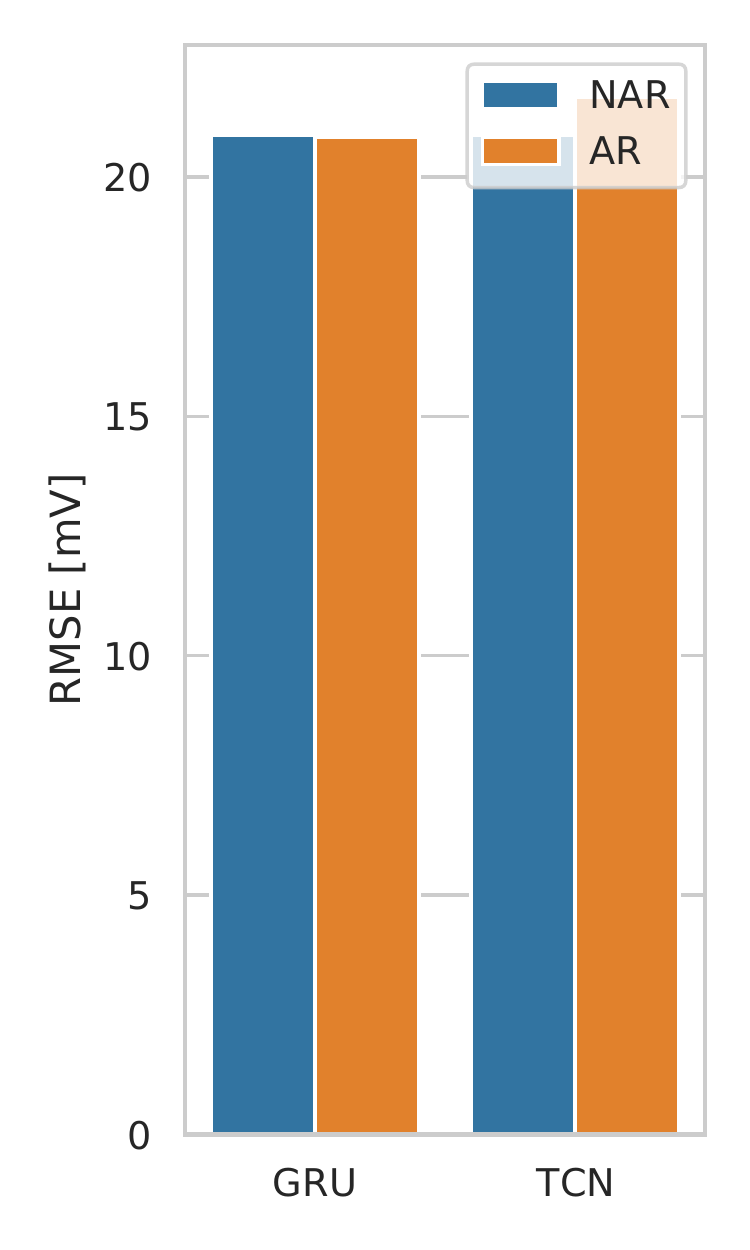}}
    \subfigure[Silverbox]{\includegraphics[width=0.15\textwidth]{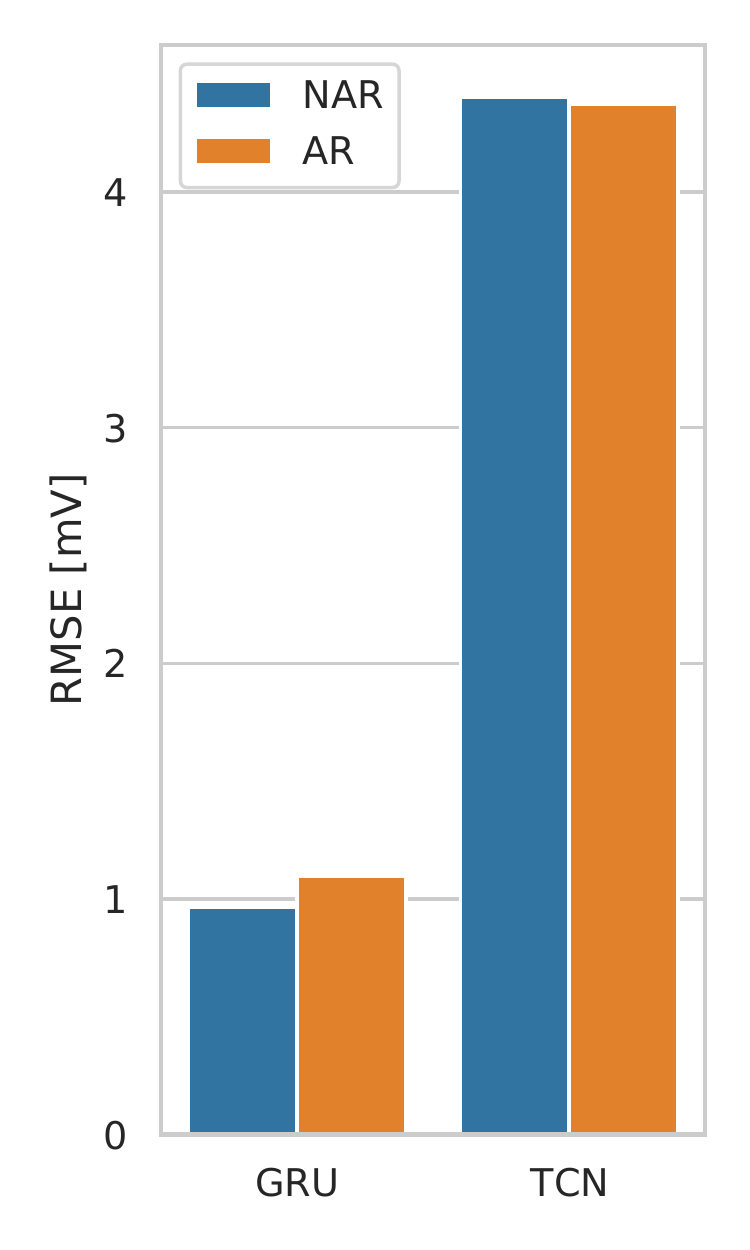}}
    \caption{RMSE of every model for the test subset of the Wiener-Hammerstein benchmark (a), the Wiener-Hammerstein benchmark with process noise (b), and the Silverbox benchmark (c). Non-autoregressive models perform better than their autoregressive counterparts.}
    \label{fig:accuracy}
\end{figure}
First, we compare the accuracy of the autoregressive and non-autoregressive variants of the models on every dataset. Figure \ref{fig:accuracy} visualizes the test RMSE of every model on every dataset. The errors of both variants are close to each other, with the non-autoregressive variant being slightly more accurate. This may be caused by a better fitting set of hyperparameters for the non-autoregressive models.

\begin{figure}
    \centering
    \includegraphics[width=0.45\textwidth]{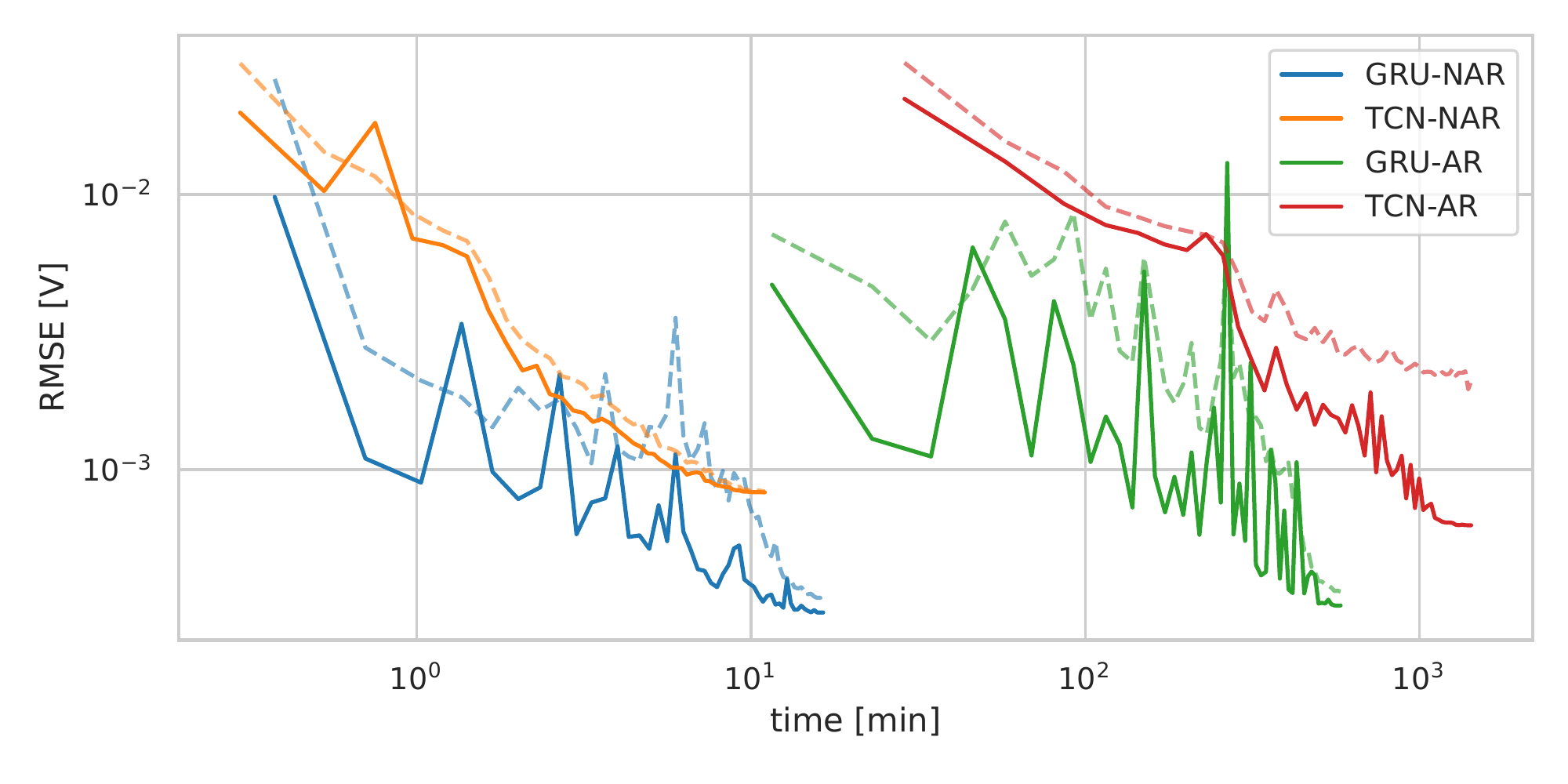}
    \caption{Validation- and training-RMSE over the duration of the training process for each model on the Silverbox dataset. The training-RMSE is drawn as dashed lines, the validation-RMSE as solid lines. The autoregressive models require significantly more time to achieve similar degrees of performance.}
    \label{fig:loss}
\end{figure}
The hyperparameter optimization process for the autoregressive models is limited by their low training speed. Figure \ref{fig:loss} visualizes the evolution of the training and validation RMSE over the time of the training of each model variant on the Silverbox dataset with the individual optimized set of hyperparameters. It becomes clear that the non-autoregressive models finish the training process significantly faster than the autoregressive models. In fact, TCN-AR is so slow, that TCN-NAR and GRU-NAR finish all mini-batches before the first mini-batch of TCN-AR is finished. This advantage in speed allows for more extensive hyperparameter optimizations as well as more complex models with the same amount of resources.

\begin{figure}
    \centering
    \includegraphics[width=0.45\textwidth]{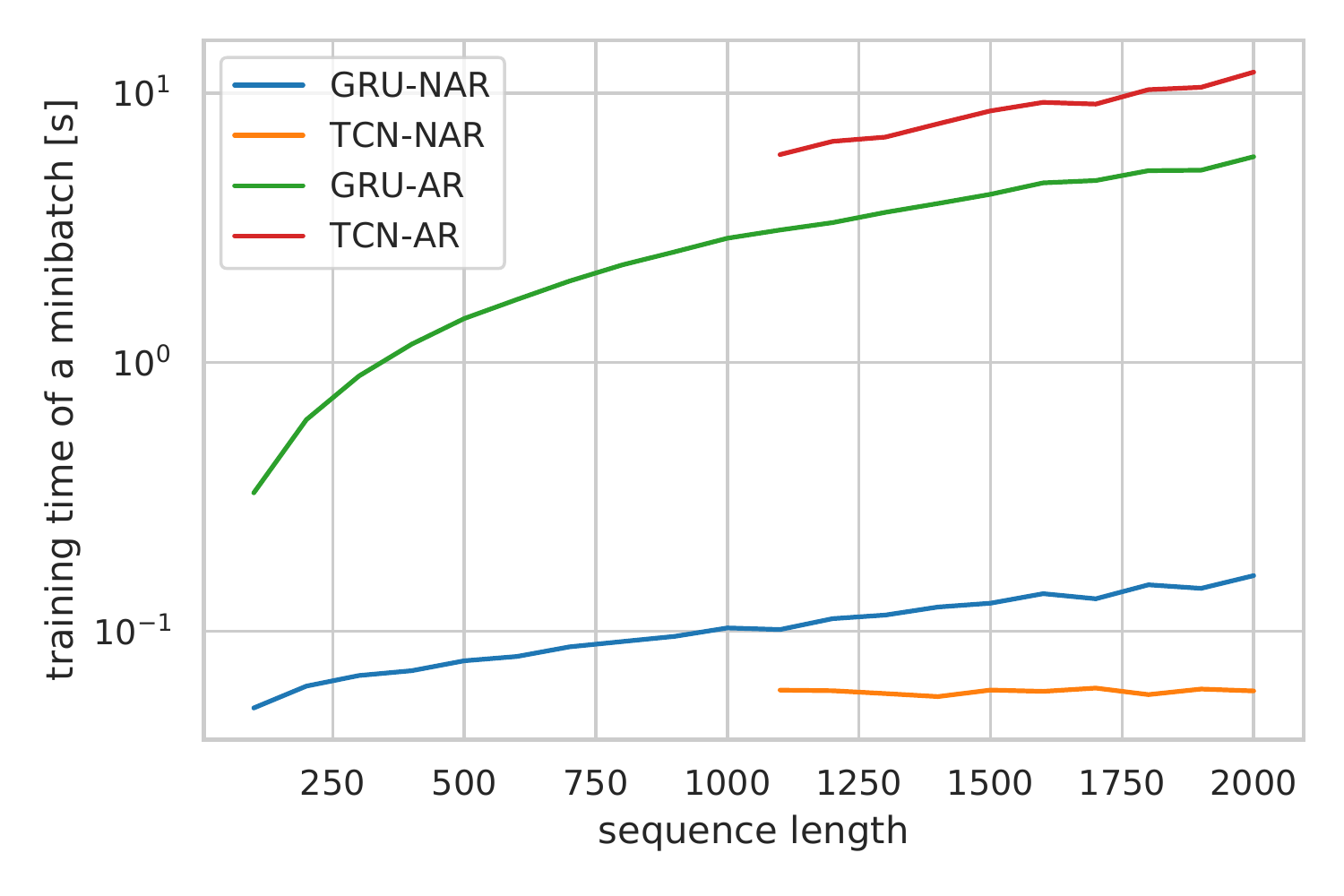}
    \caption{The training time of a mini-batch with a size of 16 over a variable sequence length. Autoregressive models scale worse than non-autoregressive models with increasing sequence lengths.}
    \label{fig:training_duration}
\end{figure}
Because of their sequential nature, the training speed of the autoregressive models is mostly dependant on the sequence length of the mini-batch. Figure \ref{fig:training_duration} visualizes the training time of a mini-batch for each model variant over the sequence length. The TCN models require a sequence length of at least 1023 samples because of the receptive field of $2^{10}-1 = 1023$ samples, which results from the depth of 10, which we identified in the hyperparameter optimization. The training time of the autoregressive models scales much worse with the sequence length than the non-autoregressive models. The training time of TCN-AR is especially bad, considering that we already use an optimized caching algorithm. In contrast, TCN-NAR has over the evaluated value range of sequence lengths a constant training duration, because the convolutions are executed independently over the sequence on the GPU. 

\begin{figure}
    \centering
    \includegraphics[width=0.45\textwidth]{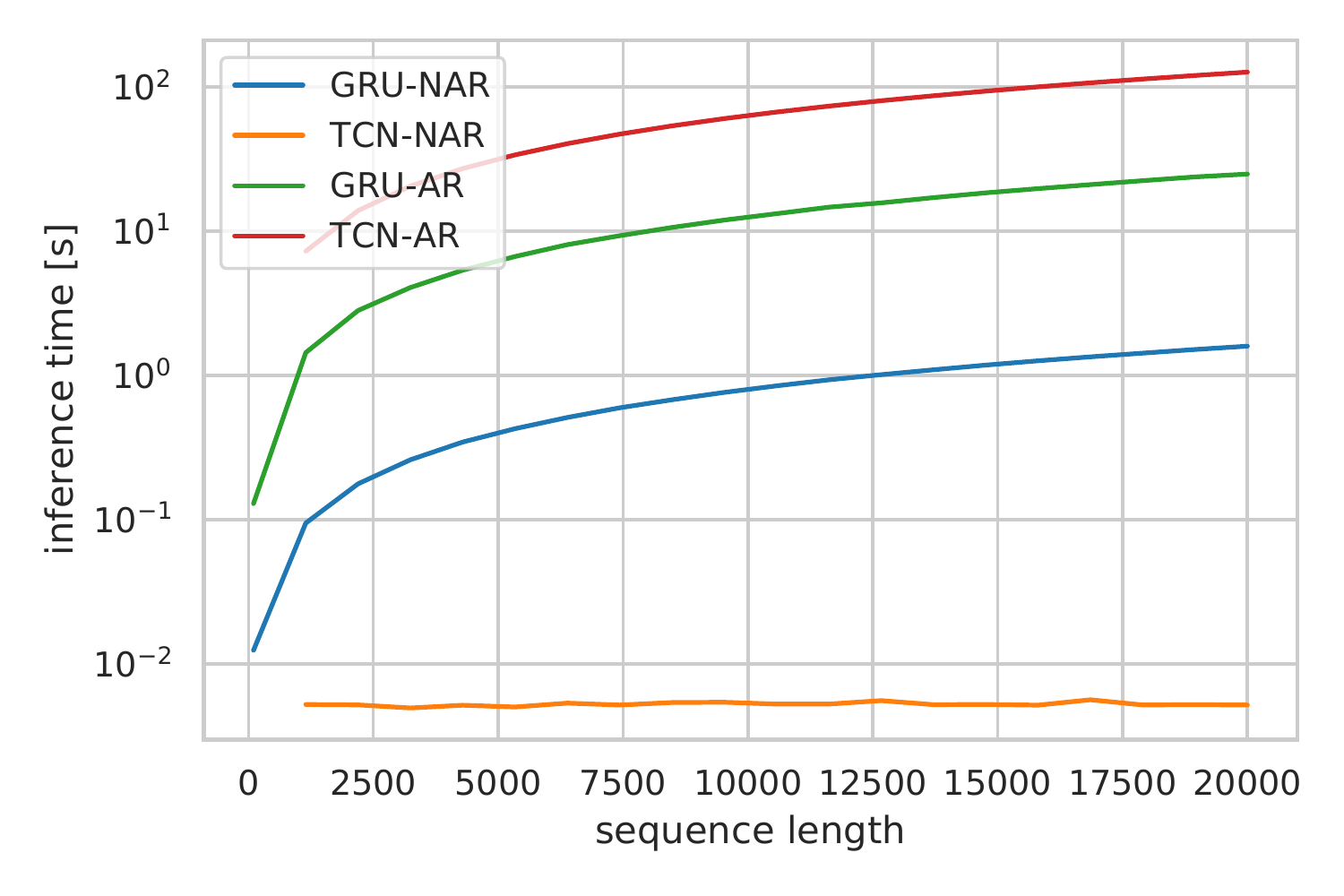}
    \caption{The inference time of a single sequence with variable sequence length for each model. Autoregressive models are slower than non-autoregressive models.}
    \label{fig:simulation_duration}
\end{figure}

The model variants also require different amounts of time for the simulation of sequences with variable lengths. Figure \ref{fig:simulation_duration} visualizes the inference time of each model variant for the simulation of sequences with variable length. The sequential models GRU-AR, GRU-NAR, and TCN-AR scale linearly with the sequence length while TCN-NAR has a constant inference time over the evaluated range because of its parallel nature. 

All in all, the autoregressive variants of the evaluated networks are slower in training and inference without any benefits for accuracy.

\subsection{Comparison with State-of-the-art Methods}
In all three benchmarks, GRU-NAR is the most accurate model of the ones, that we implemented. We compare the accuracy of GRU-NAR with the results of state-of-the-art neural network-based and other black-box system identification methods that were applied in related work on these datasets.

\begin{table}
\caption{Test RMSE of black-box models in the Silverbox benchmark}
\label{tab:res_silverbox}
\begin{tabular}{c|l}
    RMSE [mV] & Method \\
    \hline
    0.26 & PNLSS \citep{paduart_identification_2010}\\
    0.33 & NN + Cubic Regressor \citep{ljung_estimation_2004}\\
    \textbf{0.96} & \textbf{GRU-NAR (NN)} \\
    2.18 & TCN (NN) \citep{maroli_nonlinear_2019} \\
    3.98 & LSTM (NN) \citep{andersson_deep_2019} \\
    4.88 & TCN (NN) \citep{andersson_deep_2019} \\
\end{tabular}
\end{table}

\begin{table}
\caption{Test RMSE of black-box models in the Wiener-Hammerstein benchmark}
\label{tab:wh}
\begin{tabular}{c|l}
    RMSE [mV] & Method \\
    \hline
    \textbf{0.39} & \textbf{GRU-NAR (NN)} \\
    0.42 & PNLSS \citep{paduart_identification_2009} \\
    0.49 & Wiener-Hammerstein \citep{wills_estimation_2009} \\
    2.98 & BLA-based \citep{morari_identification_2010}\\
    4.71 & FS-LSSVM \citep{de_brabanter_fixed-size_2009} \\
    17.6 & CNN (NN) \citep{lopez_nonlinear_2017} \\
    45.61 & FFH (NN) \citep{romero_ugalde_neural_2013} \\
    51.4 & DN-BI (NN) \citep{rosa_nonlinear_2015} \\
\end{tabular}
\end{table}

\begin{table}
\caption{Test RMSE of black-box models in the Wiener-Hammerstein benchmark with process noise}
\label{tab:wh_noise}
\begin{tabular}{c|l}
    RMSE [mV] & Method \\
    \hline
    \textbf{20.3} & \textbf{GRU-NAR (NN)} \\
    25 & WH-EIV \citep{schoukens_identication_2016} \\
    30 & PNLSS \citep{gedon_deep_2020}\\
    30.3 &  NFIR \citep{belz_automatic_2017} \\
    42.35 & STORN (NN) \citep{gedon_deep_2020}\\
    54.1 & VAE-RNN (NN) \citep{gedon_deep_2020}\\
\end{tabular}
\end{table}
The test results are compared in Table \ref{tab:res_silverbox} for the Silverbox benchmark, Table \ref{tab:wh} for the Wiener-Hammerstein benchmark, and Table \ref{tab:wh_noise} for the Wiener-Hammerstein benchmark with process noise.

In the Silverbox benchmark, black-box system identification methods have difficulties with the test dataset because of the extrapolation part. The extrapolation performance highly depends on the estimator structure and can not generalize over all systems, which is why this is a major shortcoming of neural networks. In this case, the system has a cubic nonlinearity, which is why a combination of a neural network with a cubic regressor, and a polynomial non-linear state-space model perform better in the extrapolation part than more complex models. 

In the Wiener-Hammerstein benchmark and the Wiener-Hammerstein benchmark with process noise, GRU-NAR outperforms not only the neural network-based models but all black-box system identification models.

The comparison demonstrates, that the performance of neural network-based models depends to a large extent on the implementation of the structure and training process. This results in GRU-NAR being the best performing neural network-based implementation, which is also highly competitive with other black-box system identification methods.

\section{Conclusion}
In this work, we described the implementation of non-autoregressive and autoregressive neural networks with current best practices for system identification. We compared the accuracy, training, and inference performance of the different models on three publicly available benchmark datasets. Finally, we compared the accuracy of the best performing neural network, a non-autoregressive gated recurrent unit, with other state-of-the-art black-box system identification models.

Our results show that autoregressive neural networks require significantly more time for training and inference than their non-autoregressive counterparts, without any benefits for accuracy. This limits the extent of possible hyperparameter optimization and model capacity, especially in more complex systems. Furthermore, we found that our implementation of non-autoregressive gated recurrent units outperforms all other neural network-based system identification models in the evaluated benchmark datasets and is among the best performing black-box models. 

The present work focused on the simulation task, where no system output values are available for the model. In future work, the comparison may be done for the prediction task. While the one-step-ahead prediction task is trivial to implement non-autoregressively, the multi-step-ahead prediction is more challenging and may require a novel approach to avoid autoregression.

\bibliography{references}

\end{document}